\documentclass[10pt,twocolumn,letterpaper]{article}

\usepackage{cvpr}              

\usepackage{graphicx}
\usepackage{amsmath}
\usepackage{amssymb}
\usepackage{booktabs}
\usepackage{enumitem}
\usepackage{multirow}
\usepackage{tabularx}
\usepackage{bbding}
\usepackage{color, colortbl}
\usepackage{stfloats}

\usepackage[pagebackref,breaklinks,colorlinks]{hyperref}

\setlist[itemize]{noitemsep, nolistsep}

\usepackage[capitalize]{cleveref}
\crefname{section}{Sec.}{Secs.}
\Crefname{section}{Section}{Sections}
\Crefname{table}{Table}{Tables}
\crefname{table}{Tab.}{Tabs.}

\def\cvprPaperID{20} 
\def\confName{CVPR}
\def\confYear{2023}
\newcommand\up[1]{\textcolor{magenta}{$\uparrow{#1}$}}

\begin{document}

\title{FUTR3D: A Unified Sensor Fusion Framework for 3D Detection}

\author{Xuanyao Chen\textsuperscript{1,2,$*$} \hspace{5mm}
Tianyuan Zhang\textsuperscript{$1,3,*$} \hspace{5mm} Yue Wang\textsuperscript{5} \hspace{5mm} Yilun Wang\textsuperscript{6} \hspace{5mm} Hang Zhao\textsuperscript{1,4} \\
\textsuperscript{1}Shanghai Qi Zhi Institute \hspace{5mm}
\textsuperscript{2}Fudan University \hspace{5mm}
\textsuperscript{3}CMU \hspace{5mm} 
\textsuperscript{4}Tsinghua University \hspace{5mm} \\
\textsuperscript{5}MIT \hspace{5mm}
\textsuperscript{6}Li Auto
}

\twocolumn[{%
\renewcommand\twocolumn[1][]{#1}%
\maketitle

\begin{center}
    \centering
    \captionsetup{type=figure}
    \includegraphics[width=1.0\textwidth]{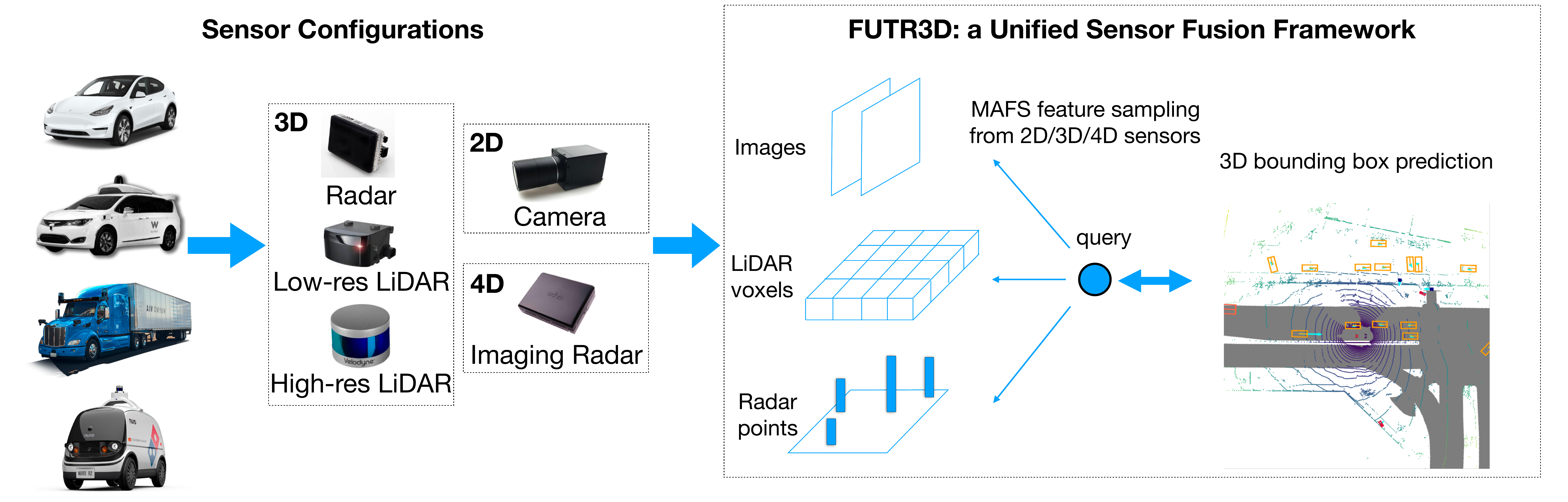}
    \captionof{figure}{Different self-driving cars have different sensor combinations and setups. FUTR3D is a unified end-to-end sensor fusion framework for 3D detection, which can be used in any sensor configuration, including 2D cameras, 3D LiDARs, 3D radars and 4D imaging radars.  } 
    \label{fig:teaser}
\end{center}%
}]

\renewcommand{\thefootnote}{\fnsymbol{footnote}} 
\footnotetext{$*$ work done at Shanghai Qi Zhi Institute.}

\begin{abstract}
   
Sensor fusion is an essential topic in many perception systems, such as autonomous driving and robotics. Existing multi-modal 3D detection models usually involve customized designs depending on the sensor combinations or setups. In this work, we propose the first unified end-to-end sensor fusion framework for 3D detection, named FUTR3D, which can be used in (almost) any sensor configuration. FUTR3D employs a query-based Modality-Agnostic Feature Sampler (MAFS), together with a transformer decoder with a set-to-set loss for 3D detection, thus avoiding using late fusion heuristics and post-processing tricks.
We validate the effectiveness of our framework on various combinations of cameras, low-resolution LiDARs, high-resolution LiDARs, and Radars. On NuScenes dataset, FUTR3D achieves better performance over specifically designed methods across different sensor combinations. Moreover,  FUTR3D achieves great flexibility with different sensor configurations and enables low-cost autonomous driving. For example, only using a 4-beam LiDAR with cameras, FUTR3D (58.0 mAP) surpasses state-of-the-art 3D detection model~\cite{centerpoint} (56.6 mAP) using a 32-beam LiDAR. Our code is available on the \href{https://tsinghua-mars-lab.github.io/futr3d/}{project page}.
\end{abstract}

\section{Introduction}


Sensor fusion is the process of integrating sensory data from disparate information sources. It is an integral part of autonomous perception systems, such as autonomous driving, internet of things, and robotics.
With the complementary information captured by different sensors, fusion helps to reduce the uncertainty of state estimation and make more comprehensive and accurate predictions. For instance, on a self-driving car, LiDARs can effectively detect and localize obstacles, while cameras are more capable of recognizing the types of obstacles. 

However, multi-sensory systems often come with diverse sensor setups. As shown in Fig.\ref{fig:teaser}, each self-driving car system has a proprietary sensor configuration and placement design. For example, a robo-taxi~\cite{wod,nuscenes,chang2019argoverse} normally has a 360-degree LiDAR and surround-view cameras on the top, together with perimeter LiDARs or Radars around the vehicle; a robo-truck usually has two or more LiDARs on the trailer head, together with long focal length cameras for long-range perception; a passenger car relies on cameras and Radars around the vehicle to perform driver assistance.
Customizing specialized algorithms for different sensor configurations requires huge engineering efforts. Therefore, designing a unified and effective sensor fusion framework is of great value.


Previous research works have proposed several sophisticated designs for LiDAR and camera fusion. Proposal-based methods either propose bounding boxes from the LiDAR point clouds and then extract corresponding features from camera images~\cite{mv3d,avod}, or propose frustums from the images and further refine bounding boxes according to point clouds in the frustums~\cite{frustumpointnet}. Feature projection-based methods~\cite{contfuse,wang2018fusing} associate modalities by either projecting point features onto the image feature maps, or painting the point clouds with colors~\cite{pointpainting}. 
Conducting camera-radar fusion usually involves more complicated feature alignment techniques due to the sparsity of Radar signals~\cite{centerfusion,rcf360,chadwick2019distant,john2019rvnet}. 


Our work introduces the first end-to-end 3D detection framework, named FUTR3D (\textbf{Fu}sion \textbf{Tr}ansformer for \textbf{3D Detection}), that can work with any sensor combinations and setups, \eg camera-LiDAR fusion, camera-radar fusion, camera-LiDAR-Radar fusion. 
FUTR3D first encodes features for each modality individually, and then employs a query-based Modality-Agnostic Feature Sampler (MAFS) that works in a unified domain and extract features from different modalities. Finally, a transformer decoder operates on a set of 3D queries and performs set predictions of objects. The design of MAFS and transformer decoder makes the model end-to-end and inherently modality agnostic.

The contributions of our work are the following:
\begin{itemize}[leftmargin=*]
    \item To the best of our knowledge, FUTR3D is the first unified sensor fusion framework that can work with any sensor configuration in an end-to-end manner. 

    \item We design a Modality-Agnostic Feature Sampler, called MAFS. It samples and aggregates features from cameras, high-resolution  LiDARs, low-resolution LiDARs and Radars. MAFS enables our method to operate on any sensors and their combinations in a modality agnostic way. This module is potentially applicable to any multi-modal use cases. 

    \item FUTR3D outperforms specifically designed fusion methods across different sensor combinations. For example, FUTR3D outperforms PointPainting~\cite{pointpainting} without bells and whistles even though PointPainting is designed to work on high-resolution LiDARs and images. 
    \item FUTR3D achieves excellent flexibility with different sensor configurations and enables low-cost perception systems for autonomous driving. On the nuScenes\cite{nuscenes} dataset, FUTR3D achieves 58.0 mAP with a 4-beam LiDAR and camera images, which surpasses the state-of-the-art 3D detection model with a 32-beam LiDAR.
    \item We will release code to promote future research. 
\end{itemize}
\vspace{-0.3em}

\section{Related Work}
\begin{figure*}[t]
    \centering
    \includegraphics[width=0.9\textwidth]{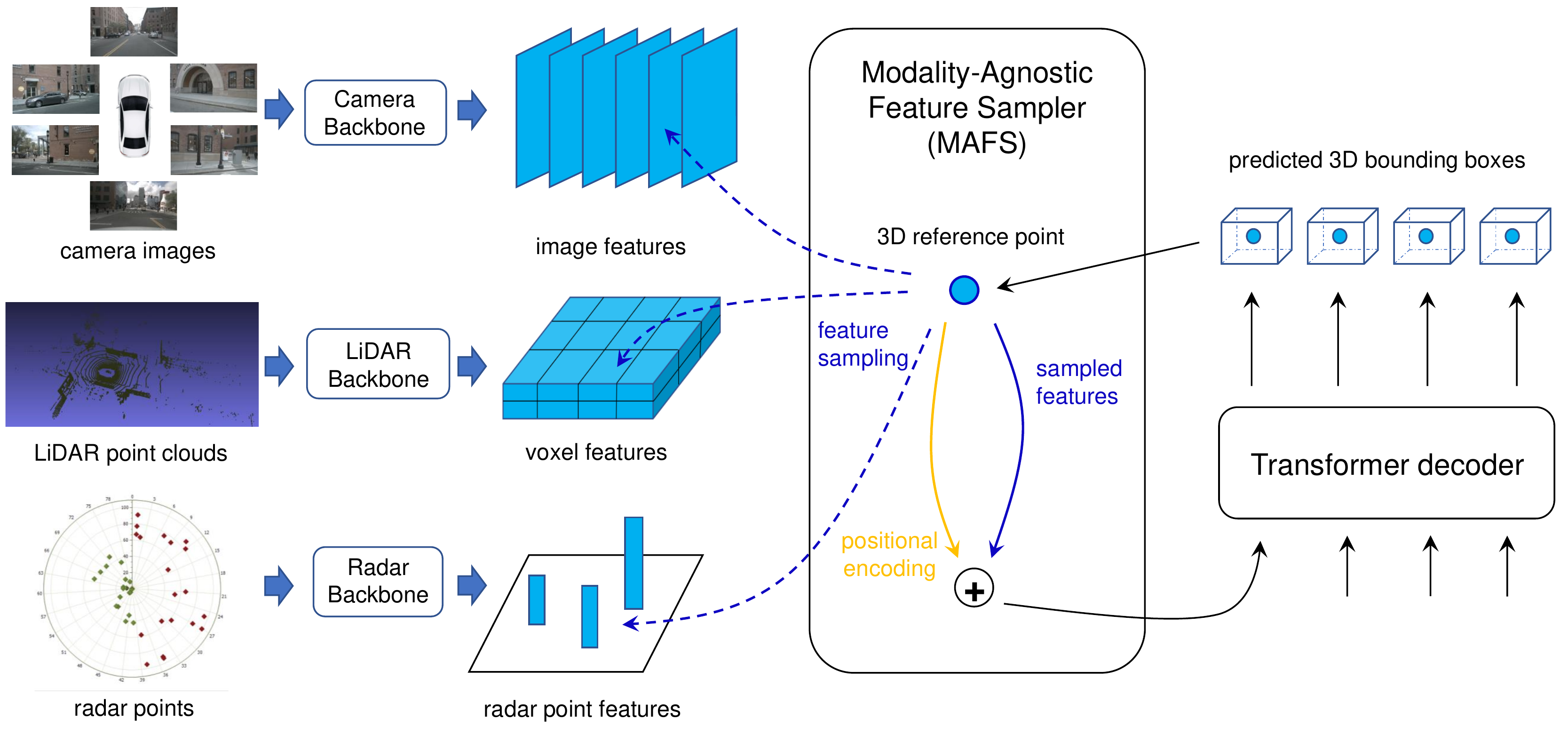}
    \caption{Overview of FUTR3D. Each sensor modality is encoded individually in its own coordinate. Then a query-based Modality-Agnostic Feature Sampler (MAFS) extracts features from all available modalities according to the 3D reference point of each query. Finally a transformer decoder predicts 3D bounding boxes from queries. The predicted boxes can be iteratively fed back into MAFS and transformer decoder to refine the predictions.}
    \label{fig:arch}
    \vspace{-1em}
\end{figure*}

\subsection{LiDAR-based 3D Detection}

The mainstreams of LiDAR-based detectors in autonomous driving quantify the 3D space into voxels or pillars, then use convolutional backbones to extract a stack of Bird's-eye view feature maps \cite{voxelnet, pointpillar, second, centerpoint}. Detectors within this framework draw lots of experiences from 2D detector designs.  Besides voxel representations,  point-based \cite{pointnet, pointnet++}, and range view \cite{velofcn ,rangedet, rsn, tothepoint} are also explored.  
PointNet architecture has been used in VoxelNet \cite{voxelnet}, Lidar-RCNN \cite{lidarrcnn} to extract feature for a small region of irregular points.  Several works \cite{rangedet, velofcn} demonstrate the computational efficiency of Range view. MVF~\cite{Zhou2019EndtoEndMF} and Pillar-OD~\cite{wang2020} introduce multi-view projection to learn view-complementary features. Object DGCNN~\cite{obj-dgcnn} models object relations using DGCNN~\cite{dgcnn} and presents the first set prediction based 3D object detection pipeline.

\subsection{Camera-based 3D Detection}
Directly migrated from 2D object detection, Monodis~\cite{monodis} learns a single-stage 3D object detector on monocular images. FCOS3D~\cite{fcos3d} considers 3D object detection on multi-view images. It predicts 3D bounding box per image and aggregates predictions in a post-processing step. Pseudo Lidar~\cite{pseudolidar} lifts images into the 3D space and employs a point cloud based pipeline to perform 3D detection. DETR3D~\cite{detr3d} designs a set-based 3D object detection model which operates on multi-view images. DETR3D uses camera transformation to link 2D feature extraction to 3D predictions. Also, it does not require post-processing, thanks to the set prediction module. Our method is closely related to DETR3D in the sense that we use a similar object detection head and feature sampling modules. In contrast to DETR3D, our feature sampling module is \textit{modality agnostic} which makes it work for sensor fusion.  


\subsection{Multi-modal 3D Detection}
Apart from classical heuristic late fusion techniques, we can roughly divide learning-based multi-modal fusion methods into two types: proposal-based methods and feature projection-based methods.

\noindent\textbf{Proposal-based} methods have gained a lot of popularity in the past few years. The idea behind such methods is to propose objects from one sensor modality, and then refine it on the other(s). MV3D~\cite{mv3d} first generates 3D object proposals in bird's eye view using LiDAR features, and then projects them to camera view and LiDAR front view to fuse LiDAR camera features. Frustum-PointNet~\cite{frustumpointnet} and Frustum-ConvNet~\cite{wang2019frustum} use 2D object detectors to generate 2D proposals in the camera view, then lift 2D proposals to 3D frustums, and finally use perform 3D box estimation within the frustums. 
AVOD\cite{avod} places dense anchor boxes in bird's eye view and then projects these anchors to camera images and LiDAR voxels for feature fusion and region proposal.

\noindent\textbf{Feature projection-based} methods usually perform feature projection and fusion before the detection head.
By finding the point-pixel correspondence, \cite{wang2018fusing} enables middle-level feature fusion, and makes a singe-stage detector. ContFuse~\cite{contfuse} further uses a KNN to better find matching points for image pixels, and fuses features at multiple levels.
The top-performing camera-LiDAR based 3D detection algorithm PointPaint~\cite{pointpainting}  projects LiDAR points onto the prediction map of a pre-trained image semantic segmentation network, and fuse the semantic prediction label with the intensity measurement of each point. MVP \cite{MVP} generate dense virtual points with semantic label to make more use of image information.
Although feature projection-based methods have recently achieved impressive performance, their designs usually require a lot of heuristics and understanding of the sensor modalities.

There are also a handful of works on fusing camera images and radar signals that share similar spirits with camera-LiDAR fusion. \cite{chadwick2019distant,john2019rvnet,rcf360} encode both camera images and radar signals in the perspective view and fuse them by simple feature map concatenation. CenterFusion~\cite{centerfusion} proposes 3D bounding boxes from images, and further refines them by fusing their features with radar signals in the bird's eye view representations.

\section{Approach}


 


FUTR3D can be conceptually divided into four parts. First, the data from different sensor modalities can be encoded by their modality-specific feature encoders (\S\ref{sec:feature-encoding}). Then, a query-based Modality Agnostic Feature Sampler (MAFS) is used to sample and aggregate features from all modalities, according to initial positions of the query (\S\ref{sec:feature-aggregate}); this is the major novelty of this work. Next, a shared transformer decoder head is used to refine the bounding box predictions based on the fused features using an iterative refine module (\S\ref{sec:iter-refine}). Finally, our loss is based on set-to-set matching between predictions and ground-truths (\S\ref{sec:loss}). FUTR3D is designed to be a unified framework for \textit{multi-modal} sensor fusion, which makes single-modal methods like DETR3D~\cite{detr3d} and Object DGCNN~\cite{obj-dgcnn} special cases of our method. For ease of presentation, we use the same notation as FUTR3D. 
An overview of FUTR3D is shown in Figure~\ref{fig:arch}.

\subsection{Modality-specific Feature Encoding}
\label{sec:feature-encoding}

FUTR3D learns features from each modality independently. Since our framework does not make assumptions about the modalities used or their model architectures, our model works with \textit{any} choices of feature encoders. This work focuses on three types of data: LiDAR point clouds, radar point clouds, and multi-view camera images. 


For LiDAR point clouds, we use a VoxelNet\cite{second,voxelnet} to encode LiDAR point clouds. After 3D backbone and FPN \cite{fpn},  we obtain multi-scale Bird's-eye view (BEV) feature maps $\{ \mathcal{F}_{\mathrm{lid}}^j \in \mathbb{R}^{C \times H_j \times W_j} \}_{j=1}^m$, where $C$ is the output channel and $H_i \times W_i$ is the size of the $i$-th BEV feature map.

We pillarize Radar points $\{ r_{j}\}_{j=1}^N \in \mathbb{R}^{C_{ri}}$ into 0.8m pillars. We adopt a MLP $\Phi_{\mathrm{rad}}$ to obtain per-pillar features $ \mathcal{F}_{\mathrm{rad}}^{j} = \Phi_{\mathrm{rad}}(r_{j}) \in \mathbb{R}^{C_{ro}}$, where $C_{ro}$ denoted the number of encoded radar features. We obtain the Radar BEV feature map $\mathcal{F}_{\mathrm{rad}} \in \mathbb{R}^{C_{ro} \times H \times W} $.

On a typical self-driving car, we have $N$ surrounding cameras. Following prior works~\cite{fcos3d,detr3d,centerpoint,pointpainting}, we use ResNet~\cite{resnet} or VoVNet~\cite{dd3d} and FPN~\cite{fpn} for image feature extraction, it outputs multi-scale features for each image, denoted as $\mathcal{F}_{\mathrm{cam}}^k = \{\mathcal{F}_{\mathrm{cam}}^{kj} \in \mathbb{R}^{C \times H_j \times W_j} \}_{j=1}^m $ for the $k$-th image.


\subsection{Modality-Agnostic Feature Sampler}
\label{sec:feature-aggregate}
The most critical part of FUTR3D is the feature sampling process, termed Modality-Agnostic Feature Sampler (MAFS). 
The input of our detection head is a set of object queries $Q=\{q_{i} \}_{i=1}^{N_q} \subset \mathbb{R}^{C}$, and features from all sensors. 
MAFS updates each query by sampling features from each sensor feature and fusing them. 

\textbf{Initial 3D reference points.} 
Following Anchor-DETR~\cite{wang2022anchor}, we first randomly initialize the queries with $N_q$ reference points $ \{c_i\}_{i=1}^{N_q}$, where $c_{ i} \in [0,1]^3$ presents relative coordinates in 3D space. Then, this 3D reference point serves as an anchor to gather features from multiple sources. The initial reference point does not depend on features from any sensors, and it will update dynamically after fusing features from all modalities. 



\textbf{LiDAR point feature sampling} 
The point cloud features after the 3D backbone and FPN~\cite{fpn} are denoted as $\{ \mathcal{F}_{lid}^j \}_{j=1}^m$. 
Following Deformable Attention~\cite{deformable_detr}, we sample $K$ points from each scale feature map.
We use $ \mathcal{P}(c_{i})$ to denote the projection of the 3D reference point in the BEV. We sample LiDAR features from all multi-scale BEV feature maps and sum them:
\begin{equation}\label{eq:lidar-weighted-sum}
    \mathcal{SF}_{\mathrm{lid}}^i = \sum_{j=1}^m \sum_{k=1}^K \mathcal{F}_{\mathrm{lid}}^j (\mathcal{P}(c_{i}) + \Delta_{\mathrm{lid}}^{ijk}) \cdot \sigma_{\mathrm{lid}}^{ijk}
\end{equation}

where $\mathcal{SF}_{\mathrm{lid}}^i$ means the sampled LiDAR point features for $i$-th reference point, $\Delta_{\mathrm{lid}}^{ijk} $ and $ \sigma_{\mathrm{lid}}^{ijk} $ is the predicted sampling offsets and attention weights, $\mathcal{F}_{\mathrm{lid}}^j (\mathcal{P}(c_{i}) + \Delta_{ijk})$ represents the bilinear sampling from the BEV feature map.

\textbf{Radar point feature sampling}
Similar to LiDAR feature sampling, we use Deformable Attention to sample Radar features as: 
\begin{equation}\label{eq:radar-weighted-sum}
        \mathcal{SF}_{\mathrm{rad} }^i = \sum_{k=1}^K \mathcal{F}_{\mathrm{rad}}( \mathcal{P}(c_{i}) + \Delta_{\mathrm{rad}}^{ik} ) \cdot \sigma_{\mathrm{rad}}^{ik}
\end{equation}

where $\Delta_{\mathrm{rad}}^{ik} $ and $ \sigma_{\mathrm{rad}}^{ik} $ is the predicted sampling offsets and attention weights.

\textbf{Image feature sampling} 
We project the reference point $c_{ i}$ to image of $k$-th camera by utilizing camera's intrinsic and extrisic parameters and denote the coordinates of projected reference point as $\mathcal{T}_k(c_i)$
We use the projected image coordinates $\mathcal{T}_k(c_i)$ to sample point features from  feature maps of all cameras, and perform weighted sum:
\begin{equation}\label{eq:img-weighted-sum}
       \mathcal{SF}_{\mathrm{cam}}^i = \sum_{k=1}^N \sum_{j=1}^m \mathcal{F}_{\mathrm{cam}}^{kj} (\mathcal{T}_k(c_{i})) \cdot \sigma_{\mathrm{cam}}^{ikj}
\end{equation}
where $\mathcal{F}_{\mathrm{cam}}^{kj} (\mathcal{T}_k(c_{i}))$ denotes the blinear sampling using the image coordinates.  Then scalar weight $\sigma_{\mathrm{cam}}^{ijk}$ is also decoded from object query $ q_i$ using a linear layer and normilzing by sigmoid. 


\textbf{Modality-agnostic feature fusion} 
After sampling point features from all modalities, we fuse features and update  queries.  First, we concatenate sampled features from all modalities and encode them using a $\mathrm{MLP}$ network $ \Phi_{\mathrm{fus}}$  given by: 
\begin{equation}\label{eq:fuse}
    \mathcal{SF}_{\mathrm{fus}}^i = \Phi_{\mathrm{fus}}(\mathcal{SF}_{\mathrm{lid}}^i \oplus  \mathcal{SF}_{\mathrm{cam}}^i \oplus \mathcal{SF}_{\mathrm{rad}}^i), 
\end{equation}
where $\mathcal{SF}_{\mathrm{fus}}^i$ is the fused per query features. Then, we add the positional encoding $ \mathrm{PE}(c_{i})$ of reference points to the fused features to make them location aware. Finally, we update the queries accordingly by $q_{ i} = q_{ i} + \Delta q_{i}$, where  
\begin{equation}\label{eq:refine}
    \Delta  q_{i} = \mathcal{SF}_{\mathrm{fus}}^i  + \mathrm{PE}(c_{i}). 
\end{equation}

Then object queries are updated using self-attention modules and FFN. Our method works with any sensor combinations thanks to this modality-agnostic feature fusion module. We use the same object detection head throughout. 



\begin{table*}[t]
\caption{\textbf{Comparison with leading methods on nuScenes \textit{test} set.} FUTR3D either surpasses or achieves comparable performance with state-of-the-art methods in single-modality settings, including those that use LiDAR and cameras separately, as well as in LiDAR-camera fusion settings. 'L' and 'C' represent LiDAR and cameras respectively. Our methods use VoVNet for camera backbones and VoxelNet with 0.075 meter size for LiDAR backbone. }

\label{tab:main}
\centering
\scalebox{0.95}{
\begin{tabular}{l| c | c c c c c c c}
\specialrule{1pt}{0pt}{1pt}

 & Modality & NDS $\uparrow$ & mAP $\uparrow$ & mATE $\downarrow$ & mASE $\downarrow$ & mAOE $\downarrow$ & mAVE $\downarrow$ & mAAE $\downarrow$ \\

\hline
FCOS3D\cite{fcos3d} & C & 40.2 & 32.6 & 74.3 & 25.9 & 44.1 & 134.1 & 16.3 \\
DD3D \cite{dd3d} & C & 47.7 & 41.8 & 57.2 & 24.9 & 36.8 & 101.4 & 12.4 \\
FUTR3D & C & 47.9 & 41.2 & 64.1 & 25.5 & 39.4 & 84.5 & 13.3\\
\hline

UVTR \cite{uvtr}& L & 69.7 & 63.9 & 30.2 & 24.6 & 35.0 & 20.7 & 12.3\\
TransFusion \cite{transfusion} & L & 70.2 & 65.5 & 25.6 & 24.0 & 35.1 & 27.8 & 12.9 \\
FUTR3D & L & 69.9 & 65.3 & 28.1 & 24.7 & 36.8 & 25.3 & 12.4\\
\hline

MVP \cite{MVP}& L+C  & 70.5 & 66.4 & 26.3 & 23.8 & 32.1 & 31.3 & 13.4 \\
FusionPainting \cite{fusionpainting}& L+C& 71.6 & 68.1 & 25.6 & 23.6 & 34.6 & 27.4 &13.2 \\
UVTR \cite{uvtr}& L+C & 71.1 & 67.1 & 30.6 & 24.5 & 35.1 & 22.5 & 12.4 \\
TransFusion \cite{transfusion}& L+C & 71.7 & 68.9 & 25.9 & 24.3 & 35.9 & 28.8 & 12.7 \\

FUTR3D & L+C  & 72.1 & 69.4 & 28.4 & 24.1 & 31.0 & 30.0 & 12.0\\

\bottomrule

\end{tabular}
}

\end{table*}

\begin{table*}[t]
\caption{\textbf{Camera with Low-Resolution LiDAR results on nuScenes \textit{val} set.} FUTR3D significantly outperforms CenterPoint with low-resolution LiDAR, and PointPainting with camera+low-resolution LiDAR.  Under camera + 4-beam LiDAR setting, FUTR3D achieves 58.0 mAP, which surpasses state-of-the-art LiDAR detector CenterPoint with 32-beam LiDAR
(56.6 mAP).}

\centering
\subfloat[
\textbf{Camera+4 beam LiDAR}
\label{tab:four-lidar}
]{
\begin{minipage}{0.48\textwidth}{\begin{center}
\scalebox{.95}{
\begin{tabular}{l|c| c c}
\specialrule{1pt}{0pt}{1pt}
 & Modality & NDS $\uparrow$ & mAP $\uparrow$\\
\hline
    CenterPoint \cite{centerpoint}& L &53.6 & 38.5\\
    FUTR3D & L & 56.4 & 44.3 \\
    PointPainting \cite{pointpainting} & L+C & 59.4 & 50.0 \\
    FUTR3D & L+C & 64.2 & 58.0 \\ 
\hline
\end{tabular}}
\end{center}}\end{minipage}
}
\hspace{-1em}
\subfloat[
\textbf{Camera+1 beam LiDAR}
\label{tab:single-lidar}
]{
\begin{minipage}{0.48\textwidth}{\begin{center}
\scalebox{.95}{
\begin{tabular}{l|c| c c}
\specialrule{1pt}{0pt}{1pt}

 & Modality & NDS $\uparrow$ & mAP $\uparrow$\\
 \hline
    CenterPoint \cite{centerpoint}& L &36.9 & 14.5\\
    FUTR3D & L & 39.2 & 16.9\\
    PointPainting  \cite{pointpainting}& L+C & 41.0 & 22.0 \\
    FUTR3D & L+C & 51.9 & 43.4 \\
\hline
\end{tabular}}
\end{center}}\end{minipage}
}
\\

\label{tab:cam-lowbeam}
\end{table*}

\vspace{0em}

\subsection {Iterative 3D Box Refinement}
\label{sec:iter-refine}

Each block $\ell\in\{1, 2, \ldots,L\}$ of transformer decoder in our detection head produces a set of updated object queries $Q^\ell=\{ q^{\ell}_i \}_{i=1}^M  \subset \mathbb{R}^{C}$.  
We predict a sequence of iteratively refined boxes given the queries. Specifically, 
for each object query $q^{\ell}_i$, we use a shared MLP $\Phi_{\mathrm{reg}}$ to predict  
offset to box center coordinate $\Delta x^{\ell}_i \in  \mathbb{R}^3$, box size ($w^{\ell}_i, h^{\ell}_i, l^{\ell}_i$), orientation ($\sin{\theta}^{\ell}_i, \cos{\theta}^{\ell}_i$), velocity ($v^{\ell}_i \in \mathbb{R}^2$) and another $\Phi_{\mathrm{cls}}$ for its categorical label $\hat{y}^{\ell}_i$. 

Following \cite{deformable_detr,detr3d,obj-dgcnn}, we adopt an iterative refinement approach. We use the predictions of box center coordinates in the last layer as the 3D reference points for each query, except for the first layer which directly decode  and input-agnostic reference points from the object queries. The next layer's reference points are given by:

\begin{equation}
    c^{\ell+1}_i = c^{\ell}_i + \Delta x^{\ell}_i
\end{equation}

    

\vspace{-2em}

\subsection{Loss}
\label{sec:loss}

Following~\cite{detr,deformable_detr,detr3d,obj-dgcnn}, we compute a set-to-set loss between predictions and ground-truths using one-to-one matching. We adopt the focal loss for classification and L1 regression loss for 3D bounding box as DETR3D~\cite{detr3d}.

As noted by Co-DETR~\cite{codetr2022}, sparse supervision, such as one-to-one set loss, can impede learning effectiveness. To learn more discriminative LiDAR feature, we incorporate an auxiliary one-stage detector head, i.e. head used in CenterPoint~\cite{centerpoint},   after the LiDAR encoder. This one-stage head is jointly trained with original transformer decoder.   Notably, the auxiliary head is only used during LiDAR training and is not used during inference.

\section{Experiments}

\subsection{Implementation Details}

\textbf{Dataset.} We use nuScenes\cite{nuscenes} dataset for all experiments. This dataset consists of 3 modalities, namely 6 cameras, 5 radars and 1 LiDAR. All are captured with a full 360-degree field of view. There are totally 1000 sequences, where each sequence has roughly 40 annotated keyframes. The keyframes are synchronized across sensors with a sampling rate of 2 FPS.

\textbf{Cameras.}
In each frame, nuScenes \cite{nuscenes} provides images from six cameras \texttt{[front\_left, front, front\_right, back\_left, back, back\_right]}; there are overlap regions between cameras and the whole scene is covered. The resolution is $1600 \times 900$. 

\textbf{LiDAR.} nuScenes provides a 32-beam LiDAR, which spins at 20 FPS. Since only key frames are annotated at 2 FPS, we follow the common practice to transform points from the past 9 frames to the current frame.  \label{para:lidar}

\textbf{Low-resolution LiDAR.} Low-resolution LiDARs are often used in many low cost uses cases. We consider these low-resolution LiDARs as complementary setups to high-resolution LiDARs because they are scalable to be deployed on production-ready platforms. We simulate low-resolution LiDAR outputs from the 32-beam LiDAR. We first convert points from Cartesian coordinate system to spherical coordinates: range $r$, inclination  $\theta$, and azimuth $\phi$. The range of inclination is [$-30^o$, $10^o$] For 4-beam LiDAR, we select beams whose inclination $\theta$ fall within $[-7.1^{\circ}, -5.8^{\circ}] \cup [-4.5^{\circ}, -3.2^{\circ}] \cup [-1.9^{\circ}, -0.6^{\circ}] \cup [0.7^{\circ}, 2.0^{\circ}]$.  For 1-beam LiDAR, we select the beam with pitch angle in $[-1.9^{\circ}, -0.6^{\circ}]$.

\textbf{Radar.}  We stack points captured by all five radars into a single point cloud; each point cloud contains 200 to 300 points each frame.  We use radar coordinates, velocity measurements, and intensities. We filter radar points using the official tool provided by nuScenes. 

\textbf{Model setting.}
The feature dimension $C$ is all 256 for LiDAR feature $\mathcal{F_{\mathrm{lid}}}$, image feature $\mathcal{F_{\mathrm{cam}}}$, object queries Q. For Radar points, $C_{ri} = 6$ and $C_{ro}=64$. There are $N_q=900$ object queries. In LiDAR and image feature extraction, we use $M=4$ layers of multi-scale features encoded by FPN. We use $K=4$ sampling offsets when using Deformable Attention. There are total $L=6$ blocks in the transformer decoder of the detection head.

\textbf{Training details.} 
For LiDAR-based detectors, we train them for 20 epochs using AdamW~\cite{adamw} optimizer. We set the learning rate as $1.0\times10^{-4}$ and adopt cyclic learning rate policy~\cite{cyclic}. We remove object sampling~\cite{second} augmentation in the last 5 epochs. For LiDAR-camera models, we pre-train the LiDAR backbone and the camera backbone respectively, then jointly fine-tune the model for another 6 epochs. For camera-radar models, we pre-train the image model, followed by joint training on cameras and radars. We set classification loss and L1 regression weights as 2.0 and 0.25. The auxiliary head loss weight is set as 0.5 in LiDAR training.

\textbf{Evaluation metrics.}
Mean Average Precision (mAP) and nuScenes Detection Score (NDS) are the major metrics for the nuScenes 3D detection benchmark. For mAP, nuScenes considers the distances between the centers of the bounding boxes on bird's eye view. NDS measures the quality of detection results by consolidating several breakdown metrics: Average Translation Error (ATE), Average Scale Error (ASE), Average Orientation Error (AOE), Average Velocity Error (AVE), and Average Attribute Error (AAE). We  average the metrics over all classes, following the official evaluation protocol. 


\subsection{Multi-modal Detection}

We demonstrate the effectiveness of our framework under several sensor combinations. 

\textbf{High-resolution LiDAR with Cameras} is the most commonly used sensor combination for autonomous driving. 
We compare our method with state-of-the-art methods in Table~\ref{tab:main}. FUTR3D achieves $72.1\%$ NDS and $69.4\%$ mAP on nuScenes \textit{test} set, surpassing TransFusion by $0.4\%$ NDS and $0.5\%$ mAP.
Moreover, FUTR3D achieves results comparable to the state-of-the-art in LiDAR-only and camera-only settings. 

\textbf{Low-resolution LiDAR with Cameras}. We also investigate the use of low-resolution LiDAR with cameras for cost-effective applications. Specifically, we simulate 4-beam and 1-beam LiDAR configurations, as described in \S\ref{para:lidar}, and compare our method against the specialized method PointPainting~\cite{pointpainting} in Table~\ref{tab:cam-lowbeam}. We report the results of the single-modality approaches with low-resolution LiDAR in Table~\ref{tab:cam-lowbeam}.

FUTR3D outperforms PointPainting by \textbf{8.0} mAP on 4-beam LiDARs plus cameras setting, and \textbf{21.4} mAP on 1-beam LiDARs plus cameras setting. PointPainting uses extra image data from nuImages to pre-train its image segmentation models while our method does not include any additional data. 
Notably, FUTR3D reaches 58.0 mAP with 4-beam LiDAR with cameras, which outperforms (by 1.4 mAP) one of the top-performing LiDAR detectors, CenterPoint \cite{centerpoint} with 32-beam LiDAR (56.6 mAP). 

FUTR3D achieves high performance with both low-resolution LiDAR and high-resolution LiDAR, which demonstrates that FUTR3D is a general sensor fusion framework.


\textbf{Cameras and Radars} is a cost-effective sensor setup commonly used in driver-assist systems for passenger cars. Radars provide sparse object localization and velocity information. In addition to the mean average precision (mAP) and normalized detection score (NDS), we also report the mean average velocity error (mAVE) metric, which measures the error of box velocity predictions and is significantly reduced by the use of radar. As demonstrated in Table~\ref{tab:camera-radar}, our method outperforms the state-of-the-art CenterFusion~\cite{centerfusion} by a significant margin. These results highlight the effectiveness of our approach in utilizing the useful information contained in the sparse radar points. Adding Radars improves over our camera-only version by 5.3 mAP and 8.6 NDS score. ResNet-101 Backbone is used for the camera backbone in this group of experiments.


\begin{table}[ht] 
\begin{center}
\caption{\textbf{Camera-Radar fusion results} on nuScenes \textit{val} set. FUTR3D outperforms CenterFusion by a large margin. Despite the sparsity of depth estimation and localization information provided by Radar points, they still contribute significantly to improving accuracy and reducing velocity error.}
\vspace{-0.8em}

\scalebox{.9}{
\begin{tabular}{l|c|ccc}
\specialrule{1pt}{0pt}{1pt}
methods & Modality & NDS $\uparrow$ & mAP $\uparrow$  & mAVE $\downarrow$ \\
\hline
CenterNet \cite{centernet}& C & 32.8 & 30.6 & 142.6 \\
CenterFusion\cite{centerfusion} & C+R & 45.3 & 33.2 & 54.0 \\
\hline
FUTR3D & C & 42.5 & 34.6 & 84.2 \\ 

FUTR3D & C+R & \textbf{51.1} & \textbf{39.9} & \textbf{41.3} \\ 
\hline

\end{tabular}
}
\label{tab:camera-radar}
\end{center}
\vspace{-2.5em}
\end{table}

\label{para:multi-modal det}

\begin{figure*}[ht]
    \centering
    \subfloat[4-beam LiDAR+Cameras vs. 32-beam LiDAR.]{
    \begin{tabular}{c}
         \includegraphics[width=2\columnwidth]{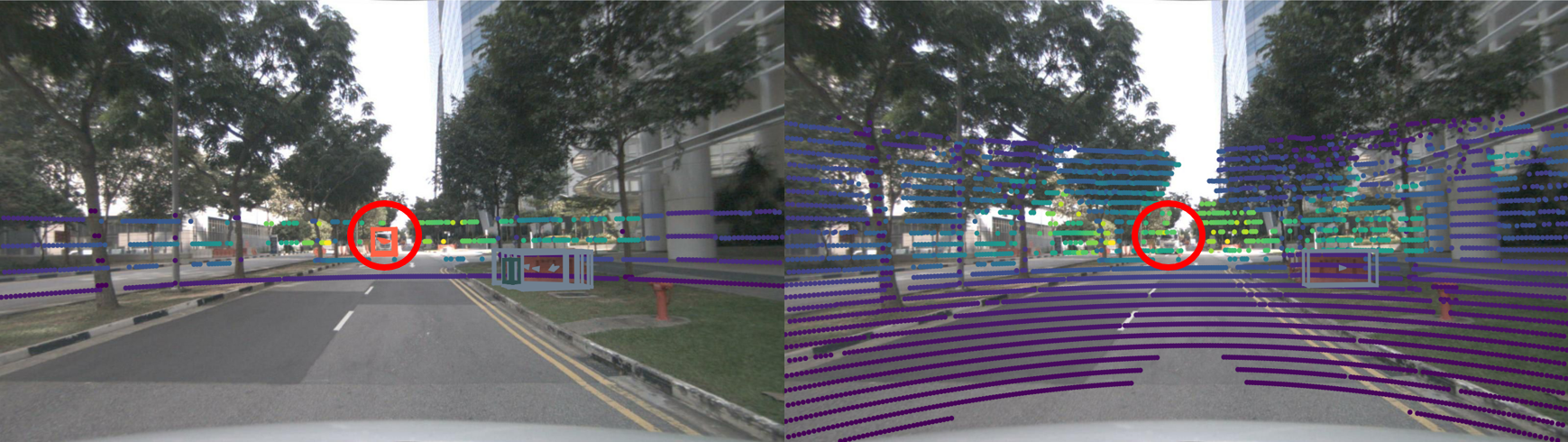}  
    \end{tabular}
    \label{fig:qua1}
    }
    \vspace{0.5em}
    \quad
    \subfloat[1-beam LiDAR+Cameras vs. Cameras.]{
    \begin{tabular}{c}
        \includegraphics[width=2\columnwidth]{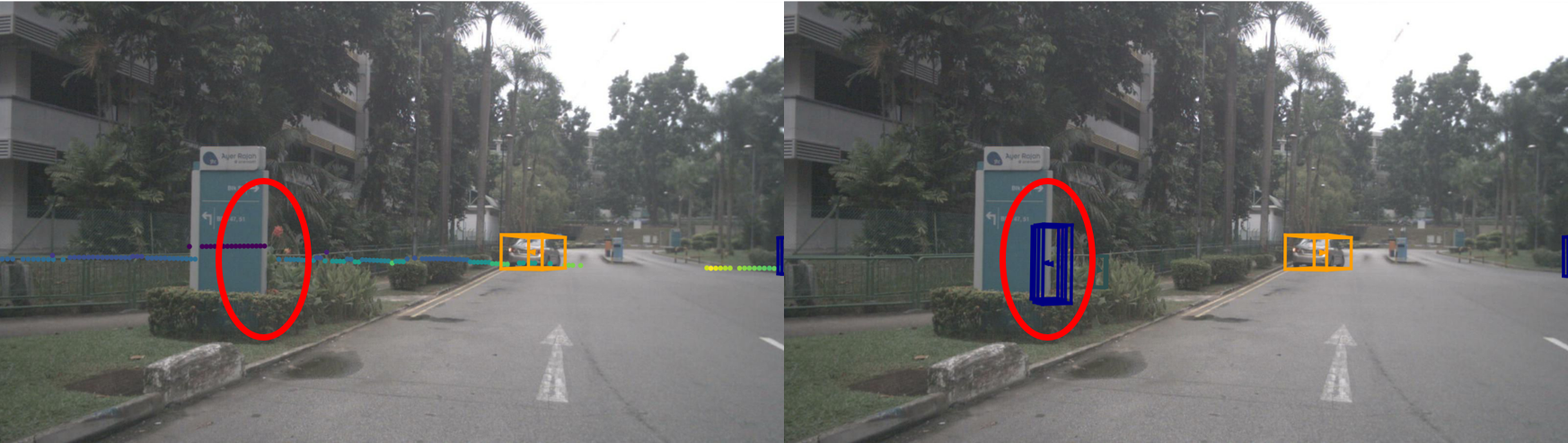} 
         
    \end{tabular}
    \label{fig:qua2}
    }
    
    \caption{Qualitative results of FUTR3D. We show perspective image view results by projecting LiDAR points onto images.
    (a) There is a car in the distance marked in red circle which are missed by 32-beam LiDAR based detector. (b) The billboard circled in red is detected falsely as pedestrian using vision only. This can be corrected with the help of 1-beam LiDAR.}
    
    \label{fig:qualitative}
    \vspace{0.5 em}
\end{figure*}

\subsection{Characteristics of Cameras and LiDARs}

\label{subsec:compare-camera-lidar}

As shown in Section \ref{para:multi-modal det}, FUTR3D is a unified detection framework that can work in camera-only, LiDAR-only and camera-LiDAR fusion settings. To the best of our knowledge, it is the first time one can control for the detection method and study the performance gain of sensor fusion comparing to each sensor respectively. To investigate the characteristics of different sensors, we break down the performance of FUTR3D on different camera-LiDAR combinations by object distances, different sizes and object categories. The resultes are based on the FUTR3D with 0.1m voxel for LiDAR and ResNet-101 for cameras.

\begin{table*}[t]
\begin{center}
\caption{Performance breakdown by object categories. Cameras help LiDAR-based detectors significantly on bicycles, traffic cones and motorcycles. Abbreviations: construction vehicle (CV), pedestrian (Ped), motorcycle (Motor), and traffic cone (TC).}
\scalebox{.9}{
\begin{tabular}{l|llllllllll}
  \specialrule{1pt}{0pt}{1pt}
  \multirow{2}{4em}{modalities} &  \multicolumn{10}{c}{AP  }\\ 
  
  & Car & Truck & Bus & Trailer &   CV  
  & Ped & Motor & Bicycle & TC & Barrier \\ \hline
  
Camera only & 54.3 & 29.8 & 36.2 & 16.7 & 7.7 & 41.6 & 32.9 & 29.6 & 51.1 & 47.6  \\

\hline

1-beam LiDAR & 29.6 & 16.0 & 30.4 & 13.8 & 3.3 & 32.5 & 6.4 & 6.0 & 9.5 & 22.0  \\
1-beam LiDAR + Camera &  61.1 & 39.7 & 49.6 & 22.6 & 12.6 & 55.2 & 39.7 & 33.1 \up{27.1} & 52.1 \up{42.6} & 47.2  \\ 

\hline

4-beam LiDAR & 70.1 & 39.4 & 50.5 & 30.0 & 12.2 & 68.6 & 44.2 & 17.3 & 41.0 & 47.5  \\
4-beam LiDAR + Camera & 78.0 & 54.4 & 61.5 & 32.2 & 20.4 & 75.7 & 61.4 & 53.9 \up{36.6} & 58.1 \up{17.1} & 53.9   \\ 

\hline
32-beam LiADR & 84.3 & 53.4 & 65.2 & 41.6 & 25.2 & 81.5 & 66.4 & 48.7 & 64.1 & 62.6 \\

32-beam LiDAR + Camera & 86.3 & 61.5 & 71.9 & 42.1 & 26.0 &  82.6 & 73.6 & 63.3 \up{14.6} & 70.1 \up{6.0} & 64.4   \\ 

 \hline
 \end{tabular}%
}
\label{tab:category}
\end{center}

\end{table*}
\textbf{Object category.} We report Average Precision (AP) of our LiDAR and cameras methods for every object category in Table~\ref{tab:category}. Though the overall mAP of the 4-beam LiDAR only FUTR3D is higher than the camera-only FUTR3D (42.1 mAP \textit{v.s.} 34.6 mAP). The camera-only model outperforms the 4-beam LiDAR model on bicycles, traffic cones and barriers, showing that 4-beam LiDAR are not good at detecting small objects. Moreover, when equipping the 4-beam LiDAR with cameras, the performance on bicycles, traffic cones and motorcycles are significantly boosted.  

\begin{table}[t]
\vspace{-0.3em }
\small
\begin{center}
\caption{Performance breakdown by object distance. We split boxes given its ego distance and report mAP independently. Results show  \textbf{cameras help LiDAR-based detectors more on farther objects.} }
\scalebox{.9}{
\begin{tabular}{l|lll}
\specialrule{1pt}{0pt}{1pt}
  \multirow{2}{4em}{} &  \multicolumn{3}{c}{mAP }\\
  & $[0m, 20m]$ & $[20m, 30m]$ & $[30m, +\infty]$\\
  \hline
    
Camera only & 49.3 & 25.4 & 10.4 \\
\hline



4-beam LiDAR & 61.1 & 39.3 & 16.1 \\

4-beam LiDAR + Camera & 69.8 \up{8.7} & 50.5 \up{10.2} & 27.4 \up{11.3} \\
\hline

32-beam LiDAR & 73.8 & 55.2 & 29.9 \\

32-beam LiDAR + Camera & 76.8 \up{3.0}  & 58.8 \up{3.6} & 36.7 \up{6.8}  \\

 \hline
 \end{tabular}
}
\label{tab:distance}
\end{center}
\vspace{-1.0em}
\end{table}
\textbf{Object distance.}
We split ground truth boxes into three subsets given the distances of box centers to ego vehicle: $[0m, 20m]$, $[20m, 30m]$, $[30m, +\infty]$, with each group taking up 42.93\%, 28.27\%, 28.8\% of all the ground truth boxes. Note that boxes of \texttt{traffic\_cone} and \texttt{barrier} with distances larger than 30 meters will be automatically filtered out following official evaluation protocols of nuScenes 3D detection. 
Table \ref{tab:distance} shows the results. For boxes farther than 30 meters, our camera-only FUTR3D only achieves 10.4 mAP, when our 4-beam LiDAR model obtains 16.1 mAP. However, fusing these two sensors elevate the performance to the next level (27.4 mAP). Even for the 32-beam LiDAR model, additional camera sensors can improve the performance of our model on farther objects from 29.9 mAP to 36.7 mAP. Adding cameras improves LiDAR perception the most on farther regions.

\begin{table}[t]
\small
\vspace{-0.3em}
\begin{center}
\caption{Performance breakdown by object size. We split the boxes given its longest edge. Results indicate that \textbf{cameras improve LiDAR-based detectors more on small objects.}}
\scalebox{.9}{
\begin{tabular}{l|ll}
  \specialrule{1pt}{0pt}{1pt}
  \multirow{2}{4em}{} &  \multicolumn{2}{c}{mAP }\\
  & $[0m, 4m]$ & $[4m, +\infty]$\\ \hline
Camera only & 22.3 & 13.9 \\

\hline

1-beam LiDAR & 9.2 & 8.9 \\
1-beam LiDAR + Camera & 25.5 \up{16.3} & 17.8 \up{8.9} \\ 

\hline

4-beam LiDAR & 25.3 & 19.4 \\
4-beam LiDAR + Camera & 33.9 \up{8.6}  & 23.4 \up{4.0} \\ 

\hline
32-beam LiADR & 36.4  & 25.7 \\

32-beam LiDAR + Camera & 39.5 \up{3.1} & 27.4 \up{1.7} \\ 

 \hline
 \end{tabular}
}
\label{tab:size}
\end{center}
\vspace{-1.0em }
\end{table}
\textbf{Object size.} We split ground truth boxes into three subsets: $[0m, 4m]$ \& $[4m, +\infty]$ based on the longer edge of the 3D box, each group occupying 46.18\%, and 53.82\% of the gt boxes. Table~\ref{tab:size} reports the mAP of our camera LiDAR models on each group.
 The performance improvements introduced by adding cameras to all LiDAR models are larger on small objects than on large objects. Cameras improve LiDAR-based detectors more on small objects since cameras have much high resolution than even 32-beam LiDARs. However, the performance improvements introduce by adding different LiDARs to camera-only models are roughly the same for small and large objects, meaning depth information are equally useful when localizing both small and large objects.

\subsection{Ablation Study}

\textbf{Auxiliary LiDAR Head.} We demonstrate the impact of our auxiliary LiDAR head on improving performance in Table~\ref{tab:aux_head}. Our results clearly show that the auxiliary LiDAR head significantly boosts performance by $3.9$ mAP. It is worth noting that the auxiliary head is only utilized during training, and model inference remains unchanged.

\begin{table}[h]
\caption{Ablation on auxiliary LiDAR Head using nuScenes validation set. The auxiliary head is only utilized in training. }
    \centering
    \begin{tabular}{c|cc}
    \specialrule{1pt}{0pt}{1pt}
   
      Aux. Head   & NDS & mAP \\
       \hline
       w/o  & 66.1 & 59.8\\
       w/  & 69.1 & 63.7 \\
        \hline
    \end{tabular}
    
    \label{tab:aux_head}
\end{table}

\textbf{Camera backbone choices.} We show the results of FUTR3D under different camera backbones. We experimented with ResNet-101 and VoVNet for image backbones. 

\begin{table}[h] 
\small
\begin{center}
\caption{Ablation on camera backbones using nuScenes \textit{test} set.  Voxelnet with voxel-size of 0.075 meter is used for LiDAR backbone.  The first row shows results of LiDAR-only version of FUTR3D.  }
\vspace{-0.5em}
\begin{tabular}{l|cc}
\specialrule{1pt}{0pt}{1pt}
Cameras & NDS & mAP \\

\hline
  --- & 69.9 & 65.3 \\
 ResNet-101 & 70.4 & 67.2 \\
VoVNet & 72.1 & 69.4\\

\hline
\end{tabular}
\label{tab:ablation-backbones}
\end{center}
\vspace{-2.0em}
\end{table}



\subsection{Qualitative Results}

In Figure~\ref{fig:qualitative}, we visualize and compare the results with different settings. In Figure~\ref{fig:qua1}, we show the results of 4-beam LiDAR + cameras (left) and 32-beam LiDAR only (right). Using sparse LiDAR beams and cameras, our method is still able to detect the car in the distance, circled in red, when it is missed using 32-beam LiDAR. Cameras provide denser pixels than LiDAR beams which could be useful for detecting far away objects. 
In Figure~\ref{fig:qua2}, we show the results of 1-beam LiDAR + cameras (left) and cameras only (right). With the help of only 1-beam LiDAR, camera is able to eliminate the false positive in red circle as LiDAR gives geometry information directly to the model. This validates the effectiveness of FUTR3D framework. Furthermore, this is in line with our assumption that in many cases, low-cost LiDARs and cameras are comparable with expensive LiDARs in object recognition. 



\section{Discussion and Conclusion}
We observe two potential limitations. First, our training pipeline requires a two-stage training: camera encoders and LiDAR encoders are first pre-trained independently on the same detection task, followed by a joint fine-tuning. This suggests a venue for further investigation into the multi-modal optimization techniques for 3D object detection. 



To conclude, in this work we propose a unified end-to-end sensor fusion framework for 3D object detection. Our insight is that a query-based modality-agnostic feature sampler (MAFS) enables models to work with any sensor combinations and setups. We hope this architecture can serve as a foundation framework for multi-modal fusion and scene understanding.  


{\small
\bibliographystyle{ieee_fullname}
\bibliography{reference}
}
\clearpage




%

\crefname{section}{Sec.}{Secs.}
\Crefname{section}{Section}{Sections}
\Crefname{table}{Table}{Tables}
\crefname{table}{Tab.}{Tabs.}
\setcounter{section}{0}
\setcounter{table}{0}
\setcounter{figure}{0}
\setcounter{equation}{0}

\def\cvprPaperID{20} 
\def\confName{CVPR}
\def\confYear{2023}


\section{Supplement}

\subsection{More visualizations}
\begin{figure*}[hb]
    \centering
    \begin{tabular}{c}
        
         \includegraphics[width=2\columnwidth]{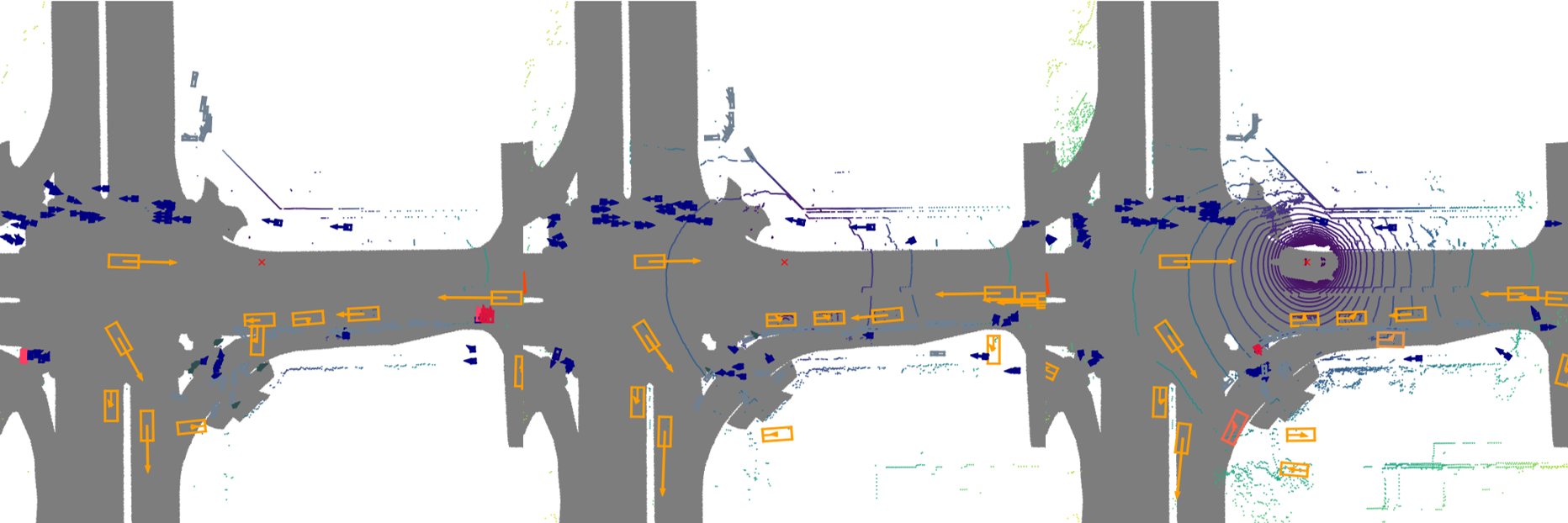}  \\
         \includegraphics[width=2\columnwidth]{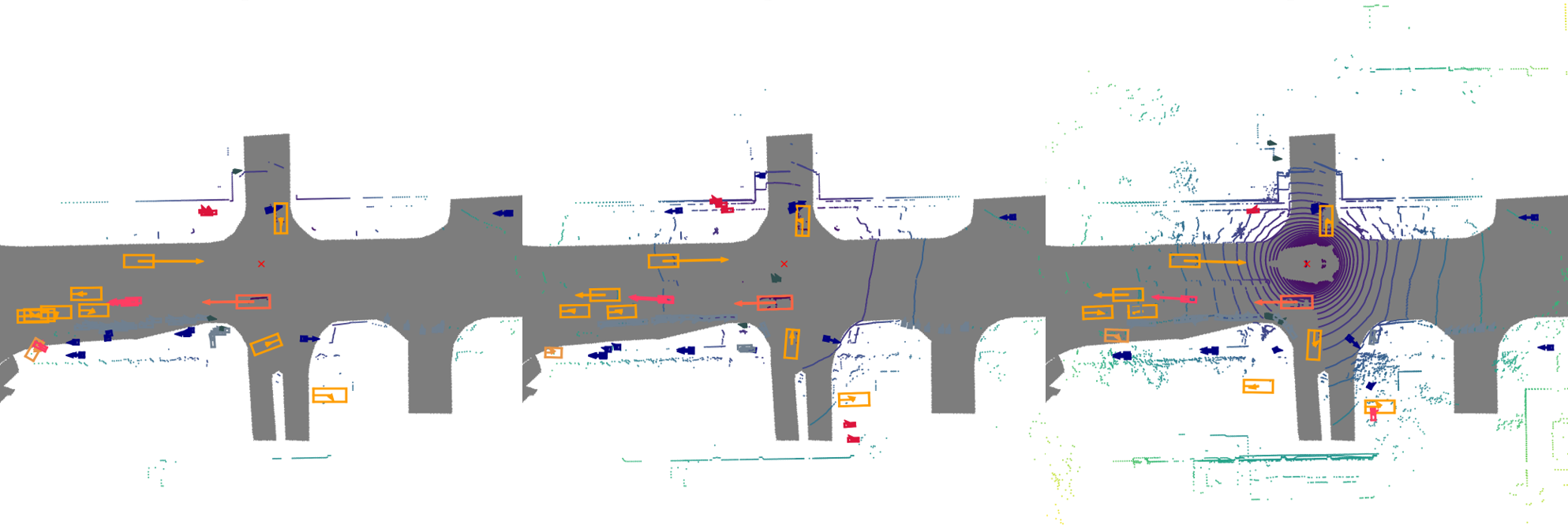}  \\
    \end{tabular}
    \caption{Results visualization in BEV. On the left is 1-beam LiDAR with cameras, in the middle is 4-beam LiDAR with cameras and on the right is 32-beam LiDAR with cameras.}
    
    \label{fig:supqua1}
\end{figure*}

\begin{figure*}[ht]
    \centering
    \begin{tabular}{c}
         \includegraphics[width=2\columnwidth]{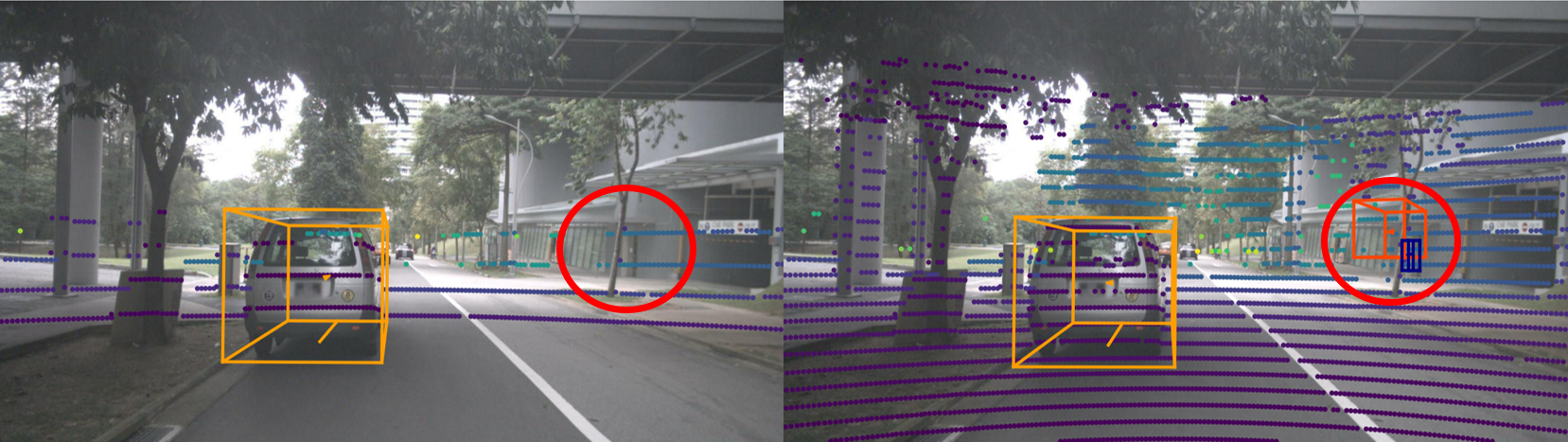}  \\
         \includegraphics[width=2\columnwidth]{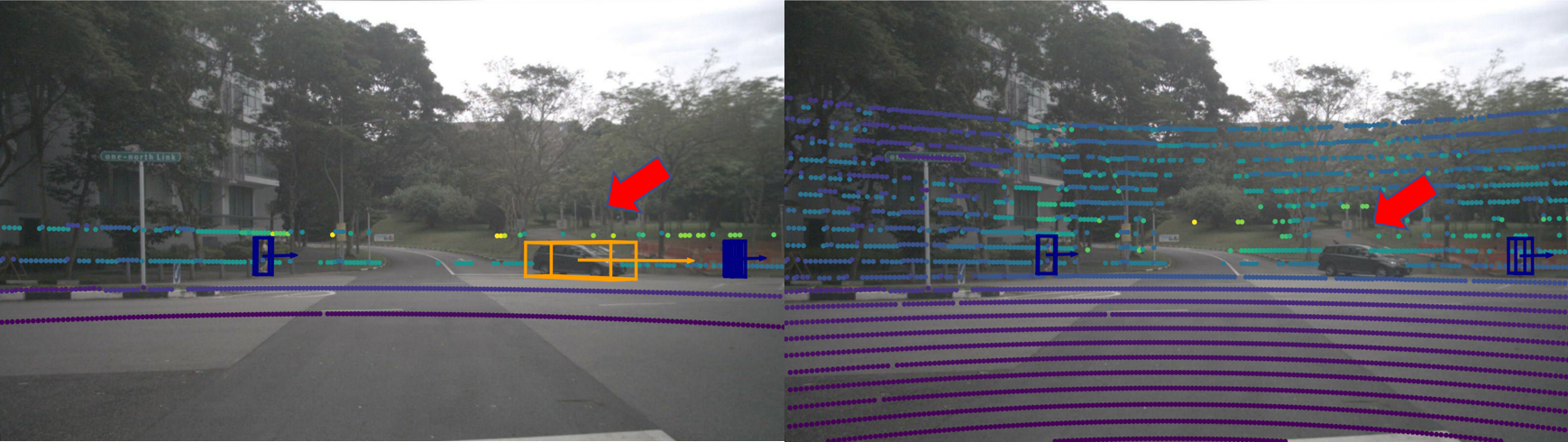}  \\
         \includegraphics[width=2\columnwidth]{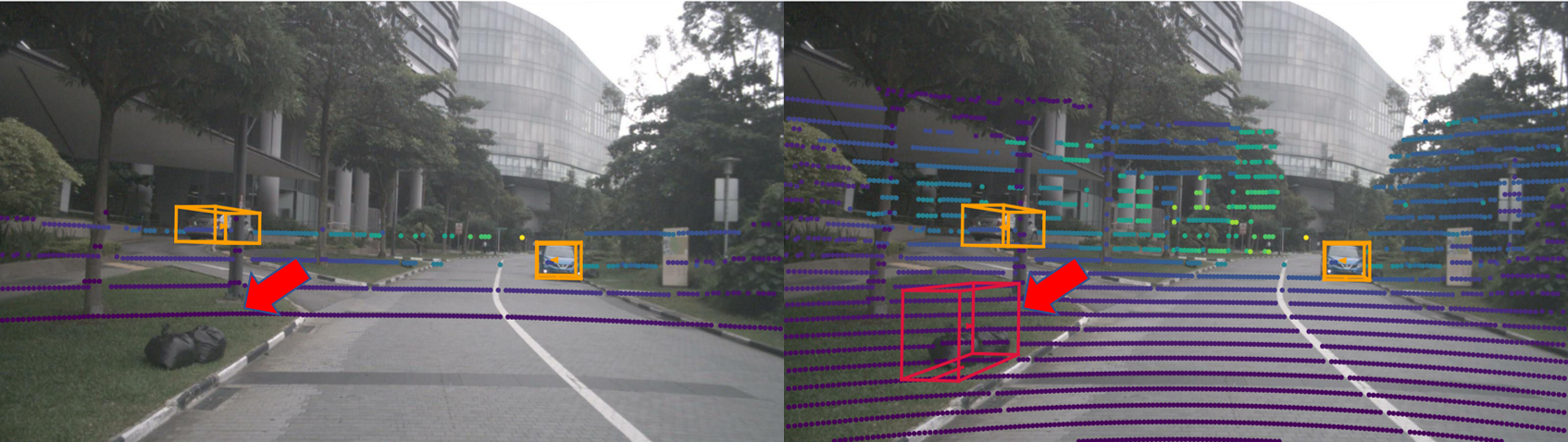}  \\
         \includegraphics[width=2\columnwidth]{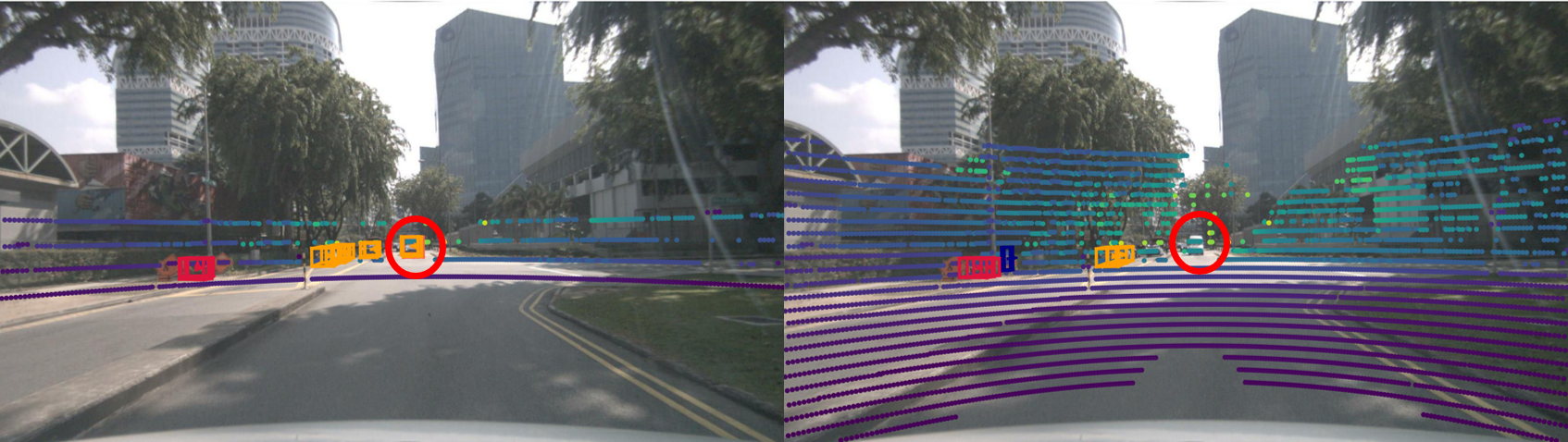}
    \end{tabular}
    
    \caption{Results visualization in perspective images. On the left is 4-beam LiDAR with cameras; on the right is 32-beam LiDAR. FUTR3D with 4-beam LiDAR with cameras achieves competitive performance compared to 32-beam LiDAR, especially for small objects like pedestrians and bicycles, and objects in the distance.}
    
    \label{fig:supqua2}
    
\end{figure*}

\begin{figure*}[ht]
    \centering
    \begin{tabular}{c}
        
         \includegraphics[width=2\columnwidth]{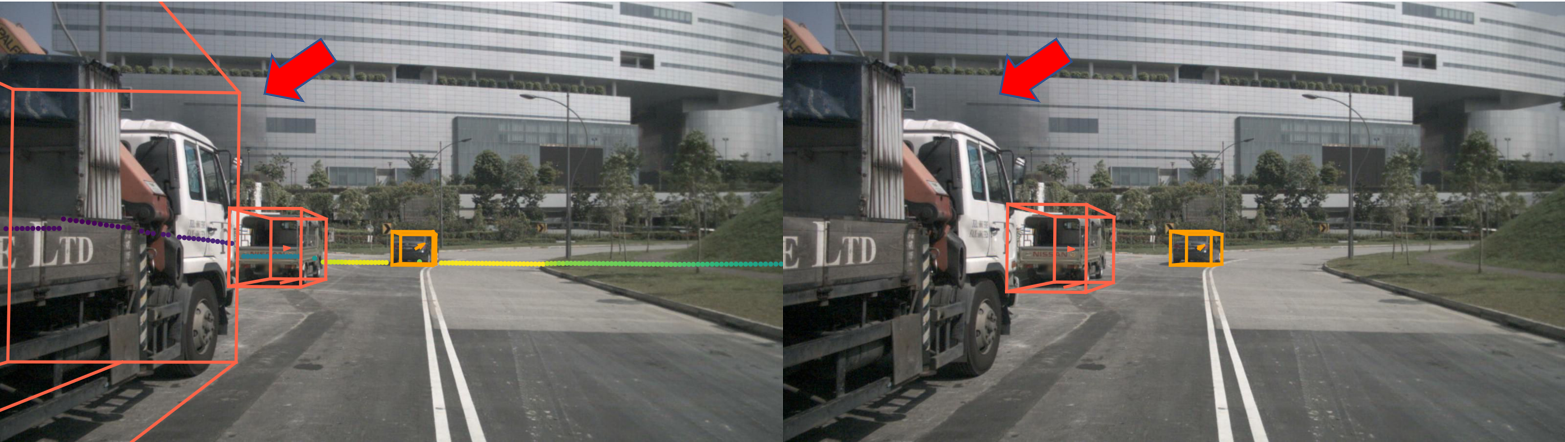}  \\
         \includegraphics[width=2\columnwidth]{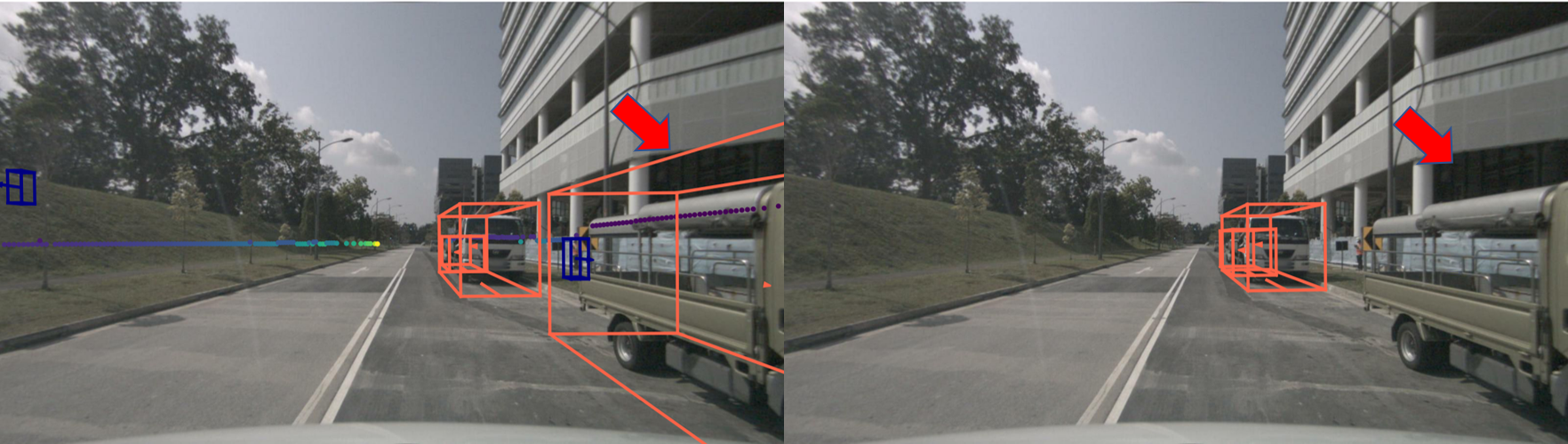}  \\
         \includegraphics[width=2\columnwidth]{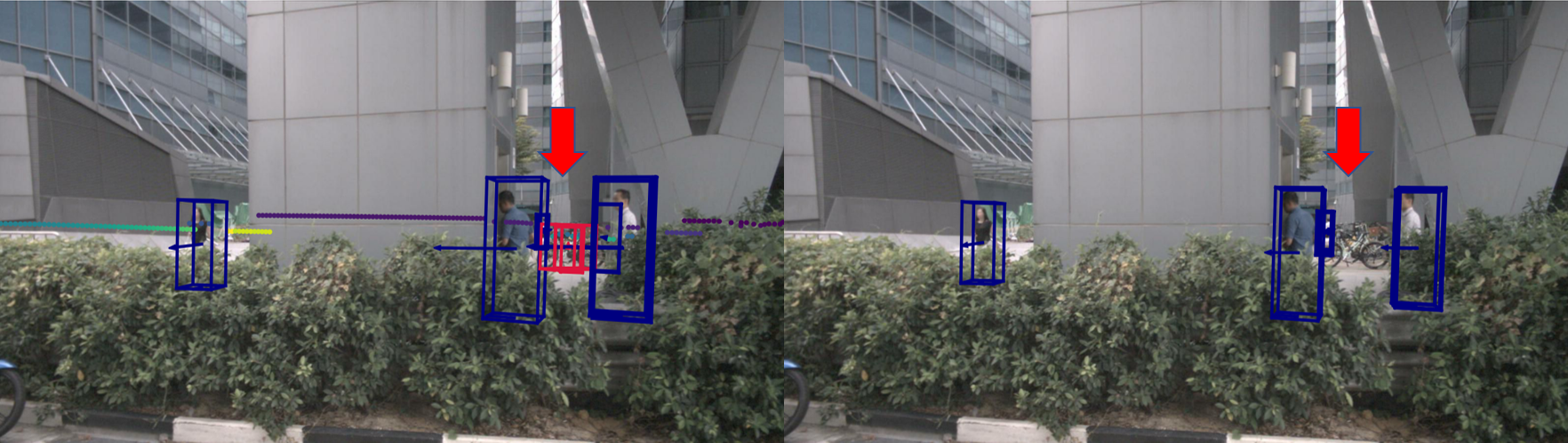}  \\
         \includegraphics[width=2\columnwidth]{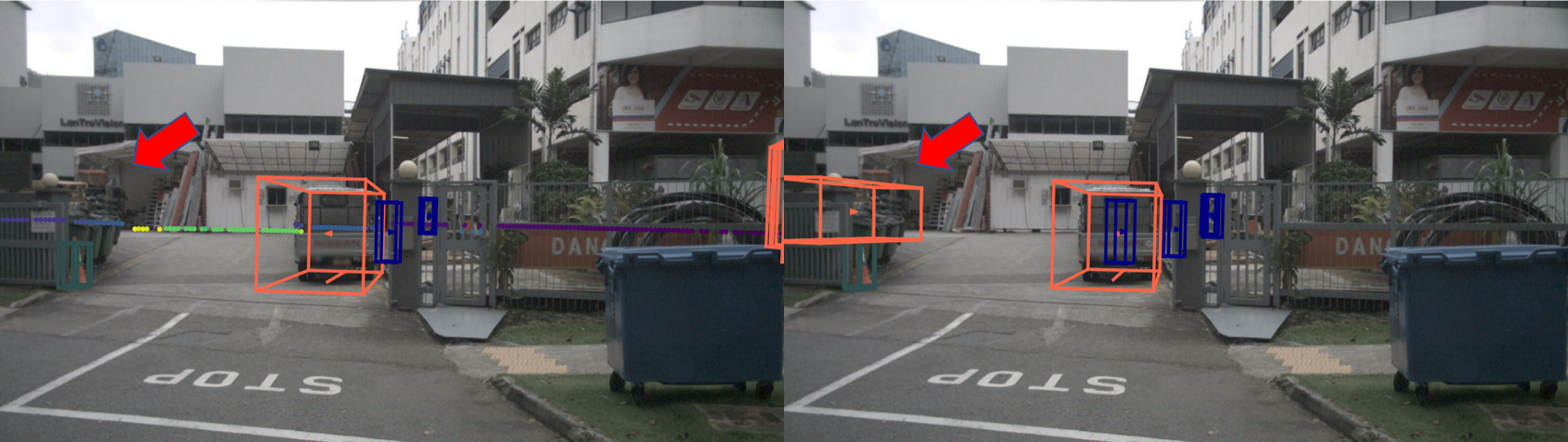} 
         
    \end{tabular}
    \caption{Results visualization in perspective images. On the left is 1-beam LiDAR with cameras, on the right is cameras. Even sparse points like 1-beam LiDAR can help FUTR3D to detect and correct the false positive.}
    
    \label{fig:supqua3}
    
\end{figure*}

We visualize the results of FUTR3D with different configurations of input modalities in Fig~\ref{fig:supqua1}, Fig~\ref{fig:supqua2} and Fig~\ref{fig:supqua3}.

\subsection{Simulation of low-resolution LiDARs}
We convert each point in point clouds from Cartesian coordinate system $(x, y, z) $ to spherical coordinates $(r, \theta, \phi)$ as

\begin{equation}\label{eq:coordinate transformation}
     \begin{split}
     r &= \sqrt{x^2 + y^2 + z^2} \\
     \theta &= \arcsin \frac{z}{\sqrt{x^2 + y^2 + z^2}} \\
     \phi &= \arctan \frac{y}{x}
     \end{split}
 \end{equation}

We pick the points according to the pitch angle $\theta$. The range of pitch angle in nuScenes is $[-30^{\circ}, 10^{\circ}]$, and the interval between adjacent beams is about $1.3^{\circ}$. For results in experiments, we select beams whose pitch angle $\theta$ fall within $[-7.1^{\circ}, -5.8^{\circ}] \cup [-4.5^{\circ}, -3.2^{\circ}] \cup [-1.9^{\circ}, -0.6^{\circ}] \cup [0.7^{\circ}, 2.0^{\circ}]$ as 4-beam LiDAR and the beam with pitch angle in $[-1.9^{\circ}, -0.6^{\circ}]$ as 1-beam LiDAR.
\subsection{Use of existing assets}

\textbf{Codebase} We use MMDetection3D \cite{mmdet3d} for baseline experiments and implementation of our own algorithms.
MMDetection3D provides a solid implementation of a wide variety of 3D detection algorithms. MMDetection3D is licensed under Apache License, Version 2.0. 

\textbf{Dataset} All of our experiments are carried out on the nuScenes \cite{nuscenes} benchmark.  The nuScenes dataset is licensed under a Creative Commons Attribution-NonCommercial-ShareAlike 4.0 International Public License (“CC BY-NC-SA 4.0”), with the additional terms listed in \url{https://www.nuscenes.org/terms-of-use}.

\end{document}